\documentclass[10pt,journal,compsoc]{IEEEtran}
\ifCLASSOPTIONcompsoc
  \usepackage[nocompress]{cite}
\else
  \usepackage{cite}
\fi

\ifCLASSINFOpdf
  \usepackage[pdftex]{graphicx}
\else
  \usepackage[dvips]{graphicx}
\fi
\usepackage{amsmath}

\usepackage{algorithmic}

\usepackage{array}

\ifCLASSOPTIONcompsoc
  \usepackage[caption=false,font=footnotesize,labelfont=sf,textfont=sf]{subfig}
\else
  \usepackage[caption=false,font=footnotesize]{subfig}
\fi

\usepackage{fixltx2e}

\usepackage{stfloats}

\usepackage{url}

\usepackage[percent]{overpic}
\usepackage{verbatim}
\usepackage{color}
\usepackage{float}
\usepackage{multirow}
\usepackage{balance}
\usepackage{soul}

\begin{document}
\title{In situ TensorView: In situ Visualization of Convolutional Neural Networks}

\author{Xinyu Chen, Qiang Guan, Li-Ta Lo, Simon Su, James Ahrens and Trilce Estrada% <-this % stops a space
\IEEEcompsocitemizethanks{
\IEEEcompsocthanksitem  Xinyu Chen and Trilce Estrada are with University of New Mexico.\protect\\
E-mail: xychen, estrada@cs.unm.edu 
\IEEEcompsocthanksitem  Qiang Guan is with Kent State University.\protect\\
E-mail: qguan@kent.edu
\IEEEcompsocthanksitem  Simon Su is with US Army Research Laboratory.\protect\\
E-mail: simon.m.su@mail.mil
\IEEEcompsocthanksitem  Li-Ta Lo and James Ahrens are with Los Alamos National Laboratory.\protect\\
 E-mail: ollie, ahrens@lanl.gov.
}% <-this % stops an unwanted space
\thanks{Manuscript received April 19, 2005; revised August 26, 2015.}}

%\markboth{IEEE Transactions on Visualization and Computer Graphics, April~2018}%
%{Shell \MakeLowercase{\textit{et al.}}: Bare Demo of IEEEtran.cls for Computer Society Journals}

\IEEEtitleabstractindextext{%
\begin{abstract}
Convolutional Neural Networks(CNNs) are  complex systems. They are trained so they can adapt their internal connections to recognize images, texts and more. It is both interesting and helpful to visualize the dynamics within such deep artificial neural networks so that people can understand how these artificial networks are learning and making predictions. In the field of scientific simulations, visualization tools like Paraview have long been utilized to provide insights and understandings. We present in situ TensorView to  visualize the training and functioning of CNNs as if they are systems of scientific simulations. In situ TensorView is a loosely coupled in situ visualization open framework that provides multiple viewers to help users to visualize and understand their networks.  It leverages the capability of co-processing from Paraview to provide real-time visualization during training and predicting phases. This avoid heavy I/O overhead for visualizing large dynamic systems.  Only a small number of lines of codes are injected in TensorFlow framework. The visualization can provide guidance to adjust the architecture of networks, or compress the pre-trained networks. We showcase visualizing the training of LeNet-5 and VGG16 using in situ TensorView.
\end{abstract}

}

\maketitle

\IEEEdisplaynontitleabstractindextext

\IEEEpeerreviewmaketitle

\begin{figure*}[h]
\centering
  \includegraphics[width=\linewidth]{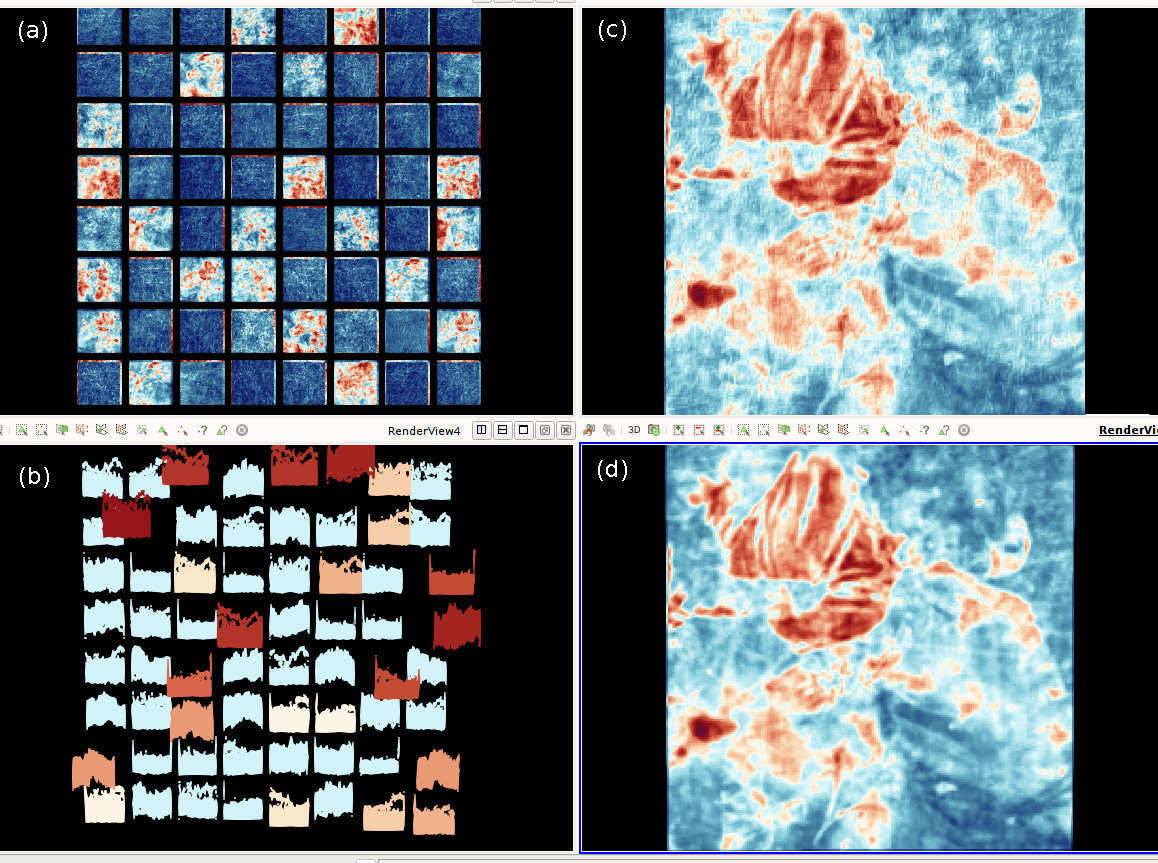}
  \caption{The first convolutional layer in the VGG16 network. (a) The activation images of one batch of samples during training. (b)The activation distributions of the same batch. Colors show similarity information. (c),(d) Two activation images show high similarity. }
	\label{fig:teaser}
\end{figure*}
\IEEEraisesectionheading{\section{Introduction}\label{sec:introduction}}
\footnotetext[1]{The
publication has been assigned the LANL identifier LA-UR-18-22809.}
\IEEEPARstart{D}{eep} Convolutional Neural Networks (CNNs) are very complex systems.  They are trained for vision  and language recognitions and many other domains such as drug discovery and genomics.\cite{lecun2015deep}. Very large CNNs, like AlexNet~\cite{krizhevsky2012imagenet}, VGGNet~\cite{Simonyan14c}, ResNet~\cite{he2016deep}, contain dozens to hundreds of layers.  Training a deep convolutional neural network is slow by two folds of challenges. The first fold is the extreme large amount of computation during the training process. The total number of trainable parameters (431k in LeNet-5, 61M in AlexNet) ~\cite{han2015learning} is big enough to understand the amount of computation that incurs.  Normally training the above mentioned convolutional neural networks need several weeks\cite{he2015convolutional}.  The second fold of challenge is the unavailable of general and efficient predefined architecture for same type of problems.  Researchers need to test different architectures and hyper-parameters such as learning rates, optimizers, batch sizes, etc. These hyper-parameters are so important that they affect the performance of networks. Sometimes, deep neural networks stop to learn because of the vanishing gradient or gradient explosion problem. Deeper and larger networks are in demand for solving modern problems. But with overly increased depth, not only the time complexity greatly increases,but also the accuracy may even saturates or drops\cite{he2015convolutional}     

To save people from trial-and-error, visualization methods are proposed to help people to understand the functions inside a large neural network.   If we have the visualization tool  to understand how neural networks update during the training and predicting phases, it will greatly help us to answer the following questions:  Do all these neurons contribute in learning and predicting?  Are there any redundant neurons or connections? How can we remove some neurons or make networks simpler? Can we find gradient vanishing or explosion problems during training process?     

Visualization techniques have been helping domain scientists to get insights from the output of scientific simulations and accelerate data-driven researches and discoveries. Widely used scientific visualization frameworks like Paraview\cite{ahrens200536} and VisIt\cite{llnlVisIt} have the following properties that satisfy the requirement of domain scientists. The first feature is they have scalability to deal with very large datasets that are produced from scientific simulations. The second feature is they enable domain scientists to do interactive exploration with their data \cite{ahrens2014image}. Consider the large computation involved in training deep neural networks and the complex searching for optimal weight parameters in high dimensional parameter spaces, Scientific visualization tools like Paraview show great potential for visualizing   the training , predicting and fine-tuning of deep convolutional neural networks.   

In this paper, we present an in situ visualization tool, in situ TensorView, which leverages the co-processing capability of Paraview Catalyst to study neural networks built upon TensorFlow's framework. In situ TensorView  is a loosely coupled In situ visualization framework\cite{Kress:2015:LCS:2828612.2828623}, which uses TensorView Catalyst adaptor to extract related data from TensorFlow then send to paraview client for In situ visualization through Catalyst pipeline. TensorView Catalyst adaptor is light-weighted with only a few lines of codes, embedded into the TensorView framework.  

We consider neural networks are similar to scientific simulation systems and the learning and predicting process are analogous to the dynamics inside a large simulation system. Data flows inside neural networks that eventually changes the connections between neurons. Such dynamics function like energy that drives particles around so we often see physics terminologies like acceleration, gradient descent, or momentum in the deep learning communities.  
Then we need to take a close look at the \emph{energy} that drives a convolutional neural networks in their learning and predicting. There are dynamics in two directions. In the feedforward pass, the input signals activate neurons layer by layer and the activations a.k.a feature maps, of each layer are finally transformed to the last fully connected layers (we would use the term \emph{activation} and \emph{feature map} interchangeably for the rest of this paper). These activations trigger some losses that go backward layer by layer to provide the gradient that we can use to adjust the connections between neurons, which are represented by the weights. Here we have the two elements that we want to visualize: activations  and weights. By visualizing the dynamics of weights, we are able see how connections are changing during training. If the learning goes wrong with gradient vanishing or explosion problems, we can quickly see connections stop changing at some points. By visualizing activations, we are able to see how neurons respond to the inputs. Then we can identify similar behaviors among neurons in the convolutional neural networks. This insight allows us to detect if some neurons are redundant. With the above qualitative and quantitative information, we can  answer the previous questions. The goal of in situ TensorView is to help researchers from various domain to build, use and understand deep neural networks  easily and comprehensively.  

The Contribution of this paper can be summarized as follows: 
\begin{itemize}
\item We introduce Paraview and visualization toolkit(VTK)\cite{schroeder2004visualization} techniques from scientific visualization to the deep learning community. These techniques can help researchers to know the status of their neural networks in training and predicting phases. To the best of our knowledge, we are the first to leverage Paraview's in situ visualization pipeline and scalability in rendering images from large dataset to visualize the training and predicting of neural networks.
\item We apply the Catalyst co-processing library to design in situ TensorView Catalyst adaptor, which is embedded in TensorFlow framework. The co-processing interface provides in situ visualization. This leverage the state of art scientific visualization techniques to deep learning problems and reduces the I/O overhead of traditional data analytics and visualizations.
\item We provide multiple viewers to show the dynamics inside neural networks. Such dynamics include statistical information about similarities between neurons and intuitive images of activations and weights at each layer. 
\item We carry out case studies to evaluate the scalability and generality of our framework. We show the visualization results both from the LeNet-5 that classifies the MNIST dataset and from the VGG16 net that classifies the 1000-class Imagenet dataset.    
\end{itemize}

The rest of the paper is organized as follows. In section 2 we discuss the related work.In section 3 we cover some background concepts about convolutional neural networks. In section 4, we explain the methodology of our framework then in section 5, we present the results in two case studies on  the LeNet-5 and the VGG16 net.  We conclude the paper in section 6.

\section{Related work}

Researchers have been working hard to understand deep convolutional neural networks. Several approaches have been proposed and they already helped people to get better understanding of the original black-box of deep learning. Such efforts falls to four categories: visualization of features, network structures, weight parameters and comprehensive frameworks. 

\subsection{Visualize class discriminant features}
The first category helps researchers to see the learned features at different layers of the neural networks.  The Deconvnet~\cite{zeiler2014visualizing} reverses the activations(a.k.a feature maps) of one layer back to an approximation of the input image. These reversed images shows the possible features in the original input spaces to demonstrate the learned features in higher layers. The visualization of activations in higher layers are obtained by reversing all the convolution, pooling, and non-linearity operations.  To reverse the max-pooling operations, the Deconvnet needs to record \emph{switches} for each input image. This is a post-analysis method and it highlights the discriminant features image by image. It cannot directly display the functioning of each convolutional filters during training or predicting phase. It dose not visualize how convolutional filters respond to a whole batch of training data.  Guided propagation ~\cite{springenberg2014striving} improves the visualization of learned features but it is also limited to  specific neural networks that have no pooling layers. 

To reverse the activations back to input spaces is generalized by \cite{yosinski2015understanding} and \cite{simonyan2013deep}. They apply the back-propagated gradients to certain regulation functions and generates expected features that have the most class discriminant power. The Deep Visualization tool box also includes an real time visualization of active neurons in a pre-trained neural network. The real-time visualization tool works with video input. This is a great step to visualize the dynamics inside a convolutional neural network during predicting phase. However it only indicates locations of active neurons on the input video frames. No statistical information is provided to help users to understand the architecture of their networks.

Methods in this category provide concrete illustration of class discriminant features. People can identify edges, corners and color patches like Gabor filters~\cite{hamamoto1998gabor}. However, they are most post-training visualizations and they focus more on the data instead of the networks per se.

\subsection{Visualize computation graphs}
Another type of visualization efforts aims to help researchers to design and examine the architectures of their deep convolutional neural networks. TensorFlow\cite{abadi2016tensorflow} help researchers to build complex networks using computation graphs to programming dozens of layers, millions of parameters and hundreds of operations. TensorFlow Graph Visualizer\cite{wongsuphasawat2018visualizing} provides the visualization of network structures so that researchers and experts do not have to read programming code to build a mental map of data flows and low-level operations. This approach works like the opposite of the aforementioned approach that visualizes the learned features.  Methods in this category are pre-analysis visualization tools for developing and designing the model structures. They do not provide information about data, training or predicting process.

\subsection{Visualize weight parameters}
The third category visualizes the weight parameters to depict searching paths through the parameter spaces. This category is less researched because of the difficulty to visualize high dimensional parameter spaces.  The approximate linear paths and the following trajectory of SGD is  studied by\cite{goodfellow2014qualitatively}. The observation provides answers to address simple qualitative questions such as \emph{Do neural networks enter and escape a series of local minima?} The learning trajectory visualized from the first 2-3 principle components (PCA) is studied by Lorch~\cite{lorch2016}. These researches try to visualize the behaviors of neural networks during training or predicting phases. Part of our work corroborate the results of the aforementioned two studies. With the support of SciVis tools, we can visualize the parameter trajectory for large network in realtime. 

\subsection{Comprehensive visualization frameworks}
There are comprehensive frameworks that provide all the above types of visualizations.  In the work of Glorot and Bengio~\cite{glorot2010understanding}, they study the changes of weights, activations, gradients, loss functions and visualize the distribution curves of these data during training phases. This is a great study to provide statistical informations about the dynamics inside neural networks. CNNVis\cite{liu2017towards} provides comprehensive and interactive visualizations of learned features, clustering of parameters and structures of neural networks. On the basis of structure graphs, CNNVis applies clustering algorithm to analyze the role of neurons of each layer according to their averaged activations. So that they can find the discriminant features in the input space for each cluster of neurons. %They explore and visualize the influence of network architectures in the case studies.  
TensorBoard ~\cite{Tensorboard} is an important comprehensive visualization framework that provide abundant visualizations to help researchers building neural networks. TensorBoard also provides  histograms to visualize loss function, weight parameters or any variables defined in TensorFlow framework.  The graph tool enables developers to visualize computation graph and operations. The Embedding Projector provides interactive visualization that users can apply t-SNE\cite{maaten2008visualizing}, PCA or user defined linear projections to transform the data of interests. These visualizations and analytics are produced from the summary data at post-training process. 

Rauber et.al use projection techniques to\cite{rauber2017visualizing} visualize the learned features at different layers and different time steps. Their research provides comprehensive analysis on network structures, data set features and neuron similarities. 
Visualizing the projection of activations enables users to qualitatively evaluate the predictive power of different network structures: The features learned by convolutional neural networks are visually more separable than multiple layer perceptrons, this corresponds that CNNs have higher accuracy. The same visualization also provides insights about the complexity of data sets. They investigate the misclassification examples and possible overfitting from the outliers and clusters in the projected visualizations. 
An important contribution of this research is they study the similarity between active neurons and the associtaiton between active neurons with certain classes, which can apply to the study of dropout and co-adaption of neurons\cite{srivastava2014dropout}. Due to the limitation of scalability in projecting high dimensional data, they most study activations in the last fully connected layer with a realtive small sample size(2000 samples).  The ActiVis\cite{kahng2018cti} by Kahng et.al is another comprehensive framework that helps researchers to design neural networks and tune hyperparameters by visualizing the activations in convolutional neural networks for both image and text data.  The framework enables users to select specific instances or subsets as input data to find relation between data, features and models. They sort neurons by their average activation values to visualize the importance of each individual neuron in predicting different classes. The ActiVis addresses the scalability challenge by using sampling(around 1000 samples ) technique and computing an activation matrix before visualization.   

Our previous work\cite{chen2017tensorview} provides the case study of visualizing the dynamics of a small CNN is also a comprehensive and post-training framework. Compared with the above three frameworks, we improve the scalability by using loosely coupled in situ visualization pipeline using VTK polygon data structure in this in situ TensorView open framework.  

\subsection{SciVis and Interactive Visualization}
Paraview~\cite{ahrens200536} is an open-source data analysis and visualization tools for scientific simulations. It provides interactive and in situ visualizations on extremely large datasets. Paraview Catalyst\cite{fabian2011paraview} is a library that integrates post-processing analysis in line with the  simulations. This integration can minimizing or eliminating data transfer bottlenecks\cite{ayachit2015paraview}. %These SciVis tools enable reseachers to formulate and test scientific hypotheses, draw conclusions and interact with data in physics, geology, climates and many other areas\cite{helbig2017challenges}. 
It is possible to leverage human's visual perception in data analytics and machine learning because machine learning methods usually obtain major improvement of solution in early iterations and only minor changes occur in later iterations. Human interactions enable users to quickly get insights and affect subsequent algorithm iterations\cite{kim2017pive}.   
%To the best of our knowledge, we are the first to leverage Paraview's in situ visualization pipeline and scalability in rendering images from large dataset to visualize the training and predicting of neural networks.
   
% \textcolor{red}{a3d-pca trajectory; qinghua clustering 2016,2017;  vast 2017 best paper.  }
 
\begin{figure}[h]
\includegraphics[width=1.0\linewidth]{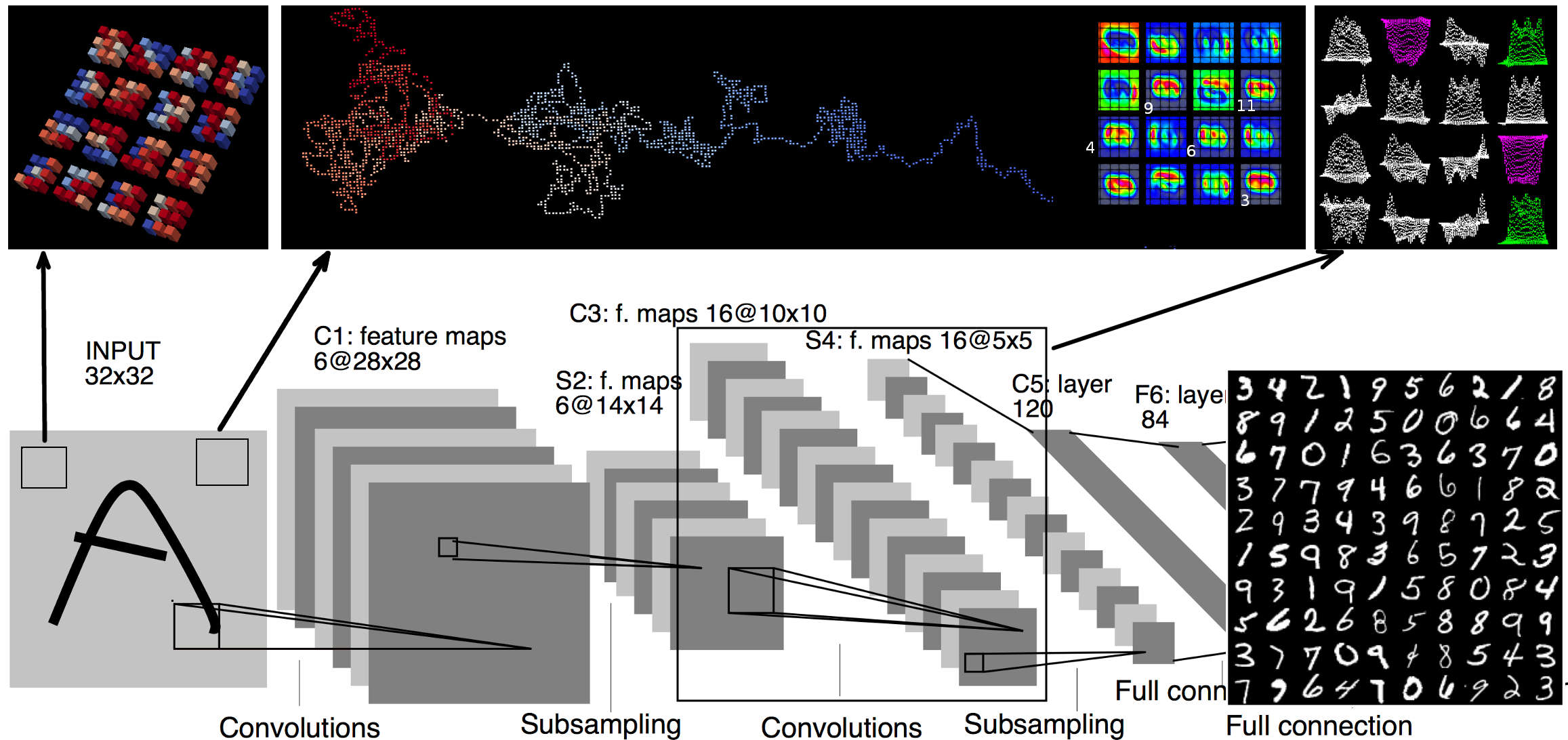}
\caption{Overview of the in situ TensorView open framework. Bottom network: our simplified LeNet-5 network with 16 and 32 filters in the convolutional layers and 256 neurons in the fully connected layer. The network is trained to classify the MNIST dataset and test our visualization methods. Top left: Visualization of the weight-grid. Top middle: Visualization the trajectory of weight parameters. Top right: Visualization of image-grid and distribution-grid.}
\label{leNet-5}
\end{figure}

\section{Background}

Artificial neural networks have massive amounts of neurons and connections. Modern deep learning frameworks often organize neurons into layers. The multilayer perceptrons(MLPs) represent such deep networks that have several fully connected \emph{hidden layers} in the middle of input and output layers.  Each neuron$_i$ in layer$_l$ gets input from neurons in previous layer$_{l-1}$ and computes wieghted sum on the input, then applies non-linear function on the sum and sends the output to next layer$_{l+1}$ .  The outputs are denoted as activations or feature maps and we use these two terms interchangeably in the rest of this paper. The activation of the above mentioned neuron$_i$ in layer$_l$ is often denoted as $a^l_{i}=\sigma(b^l_{i}+\sum_{k}w^l_{i,k}a^{l-1}_{k})$ where $\sigma$ is the non-linear function. $w^l_{i,k}$ are weight parameters and $b^l$ is bias parameter that are learned by neural networks during training.   The most widely used activation function today is Rectified Linear Units (ReLU) which is $relu(x)=max(0, x)$.

Convolutional neural networks(CNNs) consist of two other types of hidden layers: convolutional layers and pooling layers. %In the aforementioned fully connected layers, neurons connect to all neurons in adjacent layers. 
Neuron in the convolutional layer only have limited receptive field represented by windows of size $5 \times$5 or 7$\times$7 pixels. Thereby they are trained to extract local motifs that are easily detected. The convolution operations over the input images can capture the same local motif in different parts of the images\cite{lecun2015deep}. Neurons in convolutional layers are also called \emph{convolutional filters}. 
%Convolutional layers contain dozens to hundreds of convolutional filters. Thus activations are high dimensional and take large amount of storage. 
The pooling layers reduce the dimension and create an invariance to small shifts and distortions\cite{lecun2015deep}. Due to the high dimensionality and large volume of activations, the visualization of large convolutional neural networks has to deal with the challenge of scalability.

Paraview is build on the visualization toolkit(VTK) and extends VTK to support streaming of all data types and parallel execution on shared- and distributed-memory machines\cite{ahrens200536}. VTK provides the data representations for structured, unstructured, polygonal and image data and algorithms for rendering, interacting with these data.  The \emph{vtkPolyData} is the class that implements the data structure to represent vertices, lines, polygons and triangle strips\cite{schroeder2004visualization}. Scalars or vectors of attribute values of points and cells are also implemented. Therefore vtkPolyData is a flexible data structure that can represent the changes of weight parameters and activations in an artificial neural network. This enables us to use colored vertices to represent pixels in the activations or use colored 3D boxes to reprent weight values in convolutional filters. 

\section{TensorView Open Framework}
We describe the details of our open framework in this section. This includes the choice of visualization and deep learning platforms, the target data to visualize and the output and rendering of visualization. 
\subsection{The platforms}
\begin{figure}[h]
\includegraphics[width=\linewidth]{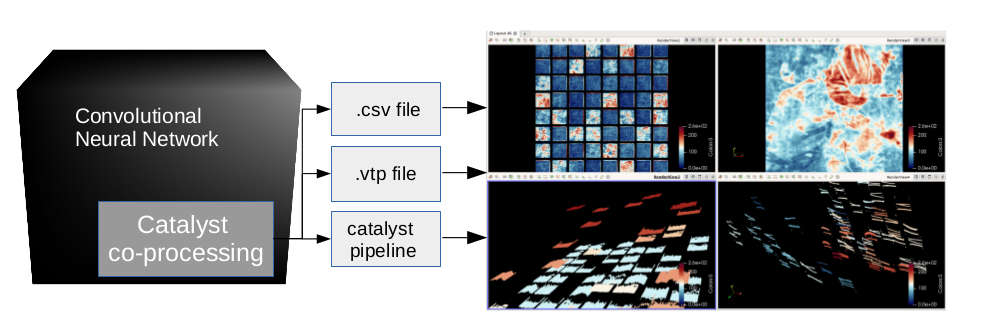}
\caption{The architecture of our open framework. Then users can choose to output .csv or .vtp file or use the Catalyst co-processing pipeline to perform in situ rendering on a Paraview client.}
\label{para-cata-arch}
\end{figure}

Our open framework leverages the power of scientific in situ visualization techniques in the deep learning problems. We choose the Paraview and TensorFlow as the building blocks of our open framework.  In situ visualization of TensorView is built on Paraview Catalyst co-processing pipeline. Figure \ref{para-cata-arch} shows the architecture of the in situ TensorView open framework. 
%The VTK polygon data structure can go through the co-processing pipeline so that Paraview can render snapshots very fast.

%In order to satisfy the scalability of visualization, we choose the scientific visualization tool, Paraview, to perform the visualization part. 
Paraview, which perform the visualization of TensorView, guarantees the scalability of visualization.  
Like large physics simulations, a deep convolutional neural network consists of large amount of parameters and produce even larger amount of data that flows through the network. For example, the first fully connected layer in the VGG16 net connects the flattened feature maps to a wide hidden layer. This turns out to have more than 100 million($25088\times4096$) parameters. This layer alone uses 406 Megabytes memory. If we want to visualize the changing of connections, we need to have the scalability to render 100 million points per time step.  Another example is the first convolutional layer of VGG16 net. The number of parameters is only $3 \times 3 \times 3 \times 64$. However, the feature maps  produced by this convolutional layer is $batchsize \times 224 \times 224 \times 64$. If we want to take a snapshot for the activations from this layer, we need to visualize $batchsize \times 3$ million points per time step.

%For the deep learning framework, we choose TensorFlow to inject our visualization tools. 
In situ TensorView can be easily embedded into TensorFlow framework.
The TensorFlow is built upon Python and a lot of the state of art networks have been implemented upon it. We provide TensorView APIs in Python. So users can directly call these APIs inside their neural network code base. Then the Paraview takes in the output and do the visualization. 

%Moreover, in situ TensorView is built on Paraview Catalyst co-processing pipeline. Figure \ref{para-cata-arch} shows the architecture of the in situ TensorView open framework. The VTK polygon data structure can go through the co-processing pipeline so that Paraview can render snapshots very fast.
\subsection{The content of visualization}
As discussed in previous sections, the most interesting data to visualize in the in situ TensorView open framework are weights and activations. During training phase, the weights are updated at each time step for many epochs. Researchers can use our visualization tool to observe the changing of weights. Normally, We expect to recognize the weights converge after some epochs. However, we will notice the weights stop changing at early stage of training if the learning rate is too high. 
%\st{During the predicting phase, or when researchers want to fine tuning a pretrained network, weight parameters in some layers  are fixed. In this case, users can bypass the visualization of weights. } 
We provide two views to visualize the changing of weight parameters. The first type is window view, which directly display the changing of weight values of convolutional filter windows. The second type is the trajectory view, which depicts the changing of weight parameters like a path through the high dimensional parameter spaces.

During training, predicting or fine-tuning phase, the visualization of activations actually help researchers to see what the networks are seeing for the moment.  We provide  two types of views to visualize activations. The first type provides statistics information that displays feature maps as if they are probability distributions. We embed the Pearson's correlation coefficient(PCC) to represent the similarity between convolutional filters into this distribution views. The second type is the image view that displays activations as pictures. Users can directly see images that resemble their training data from the this second view. 

In the following case studies, we show that combining 
%the use of 
the statistics view and the image view, researchers can get direct support to find out redundancies in their networks. 
 
\subsection{Representation and Colors}
Since Paraview is originally designed for visualizing scientific data, it can render both 2-dimensional and 3-dimensional models with various color maps to represent the data of interests. The in situ TensorView framework combines color maps and 3-dimensional models to get better representations of the weights and activations. 
          
The filters in convolutional layers are tensors that have shapes of $[w,w,c,f]$ where $w$ stands for the size of individual filter windows, $c$ refers to the number of color channels in the input, $f$ is the number of filters in the particular layer. We denote the weights of those filter windows as the \emph{weight-window}s. To visualize the weights in a convolutional layer, we flatten the last dimension of this tensor and arrange all $f$ weight-windows into a 2-dimensional grid. We denote this grid as the \emph{weight-grid}. For example, the first convolutional layer of LeNet-5 has 32 filters with weight-window of size $5 \times 5$ and one channel of gray scale. We arrange the 32 filters into a $4 \times 8$ weight-grid. Similarly in the VGG16 net,  the first convolutinonal layer can be visualized as a weight-grid of $8 \times 8 \times 3 =192$ weight-windows. Within each weight-window, we use colored blocks to represent the actual weight values.  In LeNet-5, size of weight-windows is $5 \times 5$ . Thus each of them consists of 25 blocks.  In the VGG16 net, size of weight-windows is  $3 \times 3$, then we use 9 blocks to represent each VGG16 weight-window.
  
To visualize the changing of weight-windows, we map the weight values to colors. Blue color represents negative weight and red color represent positive weight. To make the changing more obvious, we also assign the z-coordinates of these blocks equal to the weight values. Thus blocks with positive weight values also looks higher than blocks with negative weight values along z-axis.  When users use the in situ visualization methods,  sometimes the changes of color are not obvious after one time step. In this case, the position changes along the z-axis can help users to see the updates more clearly. 

In the second view to visualize the changes of weight parameters, we use a 3-dimensional model to represent the trajectory of weight parameters when networks go through the training phases. We didn't put extra analytics on the weight values but simply choose 3 arbitrary dimensions of the weight parameters as our x,y and z coordinates. Then we use colors to represent time steps. As the neural network starts the learning process, the color of a trajectory changes from blue to red. Thus we have a visualization of the searching path in a 3-dimensional parameter spaces when the weight parameters are getting updated during the training phase.

The grid layout and color maps are also used in representing the distribution view and image view of activations. We denote the visualization of distribution view and image view as \emph{distribution-grid} and \emph{image-grid} respectively. The activations in a convolutional layer are tensors that have the shape of $[b,m,m,f]$. Here we denote a batch $B$ contains $b$ samples $s_1,s_2,...,s_b$. $b$ is the batch size during training or predicting. $m$ stands for the size of feature maps and $f$ stands for the number of convolutional filters. To provide a comprehensive view of the activations, we do not visualize each sample $s \in B$. Instead, we sum the activations within a batch pixel by pixel. This gives us a tensor of shape $[m,m,f]$ and we flatten this tensor to  arrange them into grids that correspond to the previous used weight-grids. 

In the image-grids, each small window is the actual summed feature map of this batch. We denote them as the \emph{image-window}s. The size of a image-window is $m \times m$ where m is the size of feature maps.  Although users can recognize some details of their sample data from the image-windows, our experiments show it is much easier for users to recognize similar activations with the distribution-grid. The small windows in distribution-grids are the flattened vectors of the image-windows. We denote them as the \emph{distribution-windows}. The x-axis of a distribution-window represents the $m \times m$ pixel locations. The y-axis represents the activated value of those pixels. To put the the distribution-windows and image-windows into the same display scale, we normalized both their x,y coordinates to $[0,1]$.  
 
\subsection{Output and Rendering }
The in situ TensorView open framework provides flexible outputs that work for multiple data format supported by the Paraview. Thus users can use the framework according to the size and complexity of their networks.   

The most convenient method is to use ascii file for Paraview can directly import \emph{.csv} file. The advantage of using ascii file is that users can easily build the output files and use them in other visualization tools like Matplotlib. This format is good for building and testing a small neural network . The disadvantage is the scalability and storage overhead. The ascii file for the activations from the first convolutional layer of VGG16 net contains 3 million points, which takes 85 Megabytes on disk and about 60 seconds to load by Paraview. It is difficult to work with large networks.

The second  output method is the binary file that uses the VTK polygon data structure.  Users can use the Paraview or VTK Python scripts to import \emph{.vtp} file. The binary files are more efficient than ascii files to load. the feature maps mentioned above takes 35 Megabytes disk storage and only 2-3 seconds to load by Paraview. It is a good choice for large and complex neural networks and if users want to save the snapshots for later analytics.

The third output method is to use the Catalyst co-processing pipeline. This is great for in situ analysis. The data for visualization goes through the socket connection from the training server to the Paraview client.  It essentially uses the same VTK polygon data structure but entirely avoids the overhead of file I/O. Thus users are able to see the dynamics in real time. 

It is also possible to combine the above methods for different visualization requests. For example, users can visualize the changing of weight parameters at each time step by putting data into the co-processing pipeline and save snapshots files for every several time steps.  

\subsection{VTK in in situ TensorView}
\begin{figure}[h]
\includegraphics[width=\linewidth]{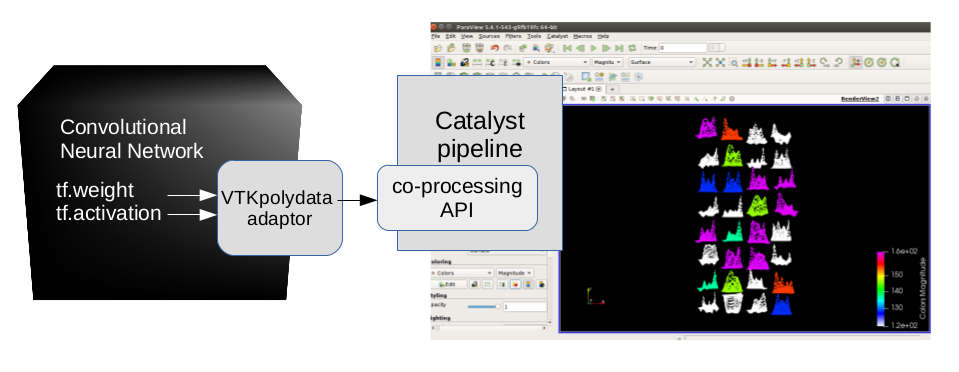}
\caption{in situ TensorView injects VTKpolydata adaptor into the neural network code base. It calls the Catalyst co-processing API to communicate with Paraview client. Users can put multiple data objects into the pipeline for in situ visualization. The Paraview also can load VTKpolydata objects from .vtp files and render them very fast.}
\label{vtk-in-tensorview}
\end{figure}

We need to convert the weight parameters and feature maps to VTK data objects to output binary files or use Paraview Catalyst co-processing pipeline. The conversion is fulfilled by injecting small amount of codes(less than 20 lines of codes for one VTKpolydata object) into the neural network code bases. Figure\ref{vtk-in-tensorview} shows the Paraview Catalyst pipeline that connects the neural network and the Paraview client. Each data adaptor extracts data from TensorFlow and puts them into the VTK polygon data structures. 

\section{Case Studies}
In this section, we carry out two case studies to illustrate how users can use our open framework to visualize their neural networks. In the first case study, we choose a simplified LeNet-5 that are trained to learn the MNIST dataset (Fig.~\ref{leNet-5}). This is a small convolutional neural network that we can train from scratch and test our methods relatively quickly. In the second case study, we choose the VGG16 net. This is a large network that can recognize 1000 classes for the Imagenet dataset. We take pretrained weight parameters to test our visualization methods. This case study is  helpful for users that want to use transfer learning and fine-tune their own networks.   

To illustrate our methods, we only present the visualization of weights and activations from the first convolutional layer. This is not because of the limitations of in situ TensorView framework. We can fulfill the visualization of all the convolutional layers with only small amount of fix in the networks.  We consider to present the visualization of the first convolutional layer for two reasons. First, although the number of weight parameters in the first convolutional layer are relative smaller than the following layers, the activations a.k.a feature maps produced in this layer is very large. The feature maps normally have the same size as the input images in this layer. On the other hand, feature maps in the following layers are pooled and shrunk to smaller sizes. Thus the visualization of the first convolutional layer requires us to consider the output format and scalability of rendering. The second reason is this layer is at the end of back-propagation path. The weight parameters here are the ones that are more prone to get stuck if the learning is affected by gradient vanishing or gradient explosion issues. We are then more interested to observe the problems at the first convolutional layer. 

We use Paraview 5.2.0 and Catalyst co-processing libraries for the visualization tasks. We use Python Matplotlib to illustrate the similarity information we put into the statistics-grid. However, our open framework works well without calling any other visualization packages such as Matplotlib. The convolutional networks are built on the TensorFlow framework. Experiments were carried out on machines with Nvidia P100 GPU cards.\\

\subsection{Simplified LeNet-5}
Our simple LeNet-5 has 16 filters in convolutional layer$_1$ and 32 filters in convolutional layer$_2$. The filter size is $3\times3$. We even reduced the number of neurons in the fully connected layer to 512. The simplification can help us to speed up our visualization experiments and it acts as a proof of concept that we can find out redundant filters. Although this network is about half the size of the original LeNet-5, in situ TensorView still picks out some similar filters from this simplified CNN. After remove the redundant filters, we achieves 99.17\% accuracy after 40 epochs of training. 
\begin{figure}
\includegraphics[width=0.95\linewidth]{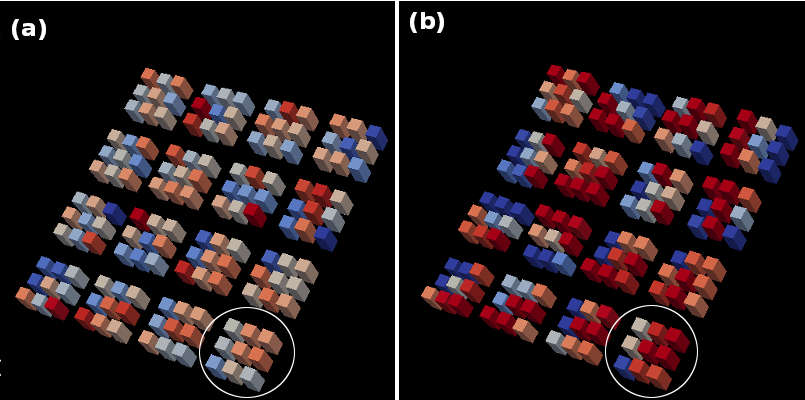}
\caption{The weight-grid$_1$ in simplified LeNet-5. Red blocks are positive weights. Blue blocks are negative weights. (a)The weight-grid at initial state. (b)The weight-grid after 8 epochs of training. The circled weight-window represents convolutional filter$_3$. The index of filters begins at zero, and starts from the lower left corner of weight-grids. }
\label{cnv1filters}
\end{figure}
\subsubsection{weight-grid}
Fig.~\ref{cnv1filters} shows the weight-grid of convolutional layer$_1$. The 16 weight-windows are arranged into this $4\times4$ grid.  Paraview can import time-series data from ascii files and binary files. With Catalyst co-processing library, Paraview can directly render the weight-grid from the in situ pipeline. Thus, we can display the changing of weight-grid in real time. In cases when learning rates are too high, it is easy to see that the colored blocks get stuck early during the training phase. The index of filters starts from zero. The filter$_0$ locates at the lower left corner of weight-grids. This grid layout is same for the distribution-grids and image-grids in the visualization of activations. The circled weight-windows represent filter$_3$ at two time steps. The left weight-grid(a) has lighter colors. It represents the initial state of this network. The right weight-grid(b) has darker colors. It represents the updated weights after 8 epochs of training. Users can observe the elevations of colored blocks have changed after some time steps.    

\subsubsection{Visualizing learning trajectories}

\begin{figure*}[t]
\includegraphics[width=1.0\linewidth]{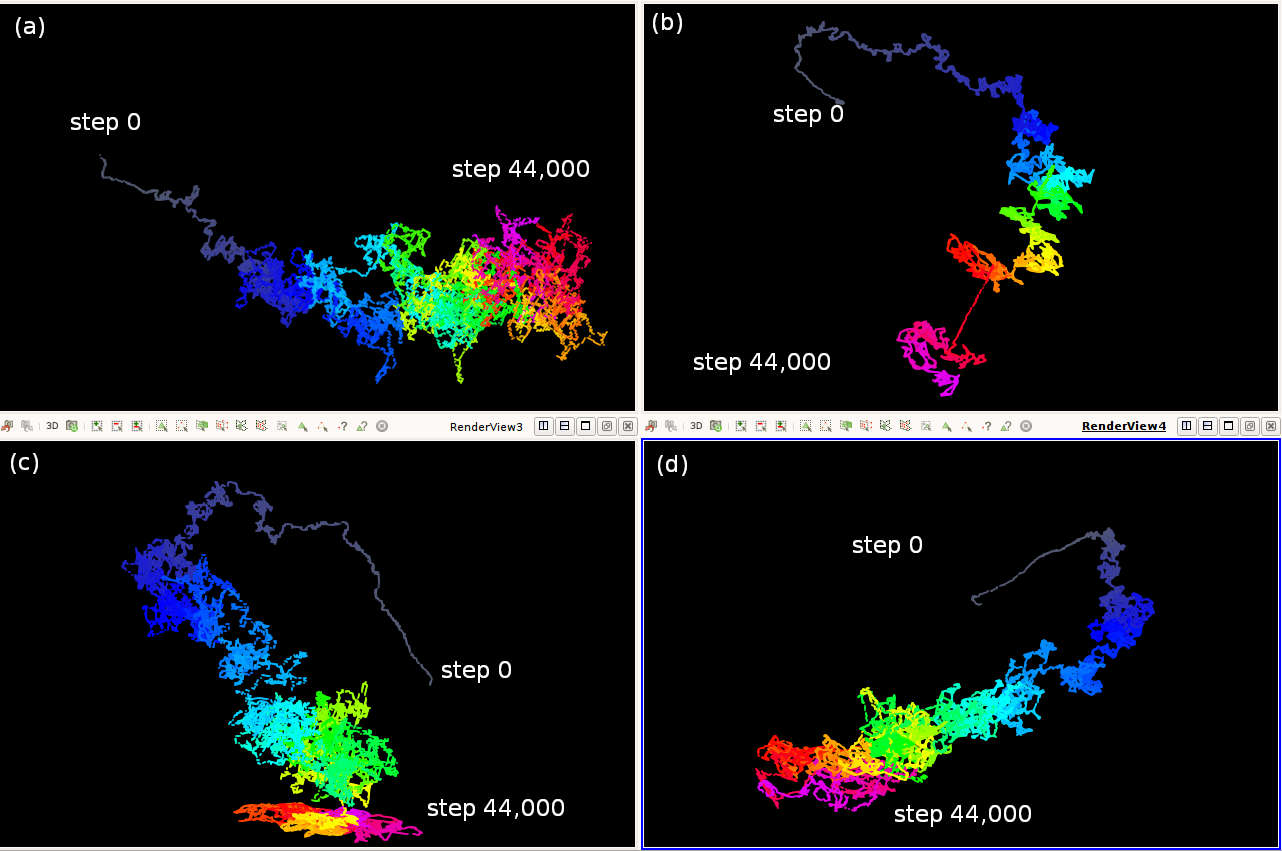}
\caption{Visualization of 4 trajectories of the weights in the first convolutional layer of simplified LeNet-5. They use different dimensions of the weight parameter space. Color represents time steps, starts from $0$ as gray to $44,000$ as pink. (a)visualize the searching path in the space of the \{0,1,2\} dimensions of the weight parameters. (b)is in the space of the \{3,4,5\} dimensions. (c)is in the space of the \{6,7,8\} weight dimensions. (d)is in the space of the \{100,101,102\} weight dimensions.}
\label{rwk4}
\end{figure*}
From the weight-grid in Figure~\ref{cnv1filters}, users can observe that the blocks that starts from blue always changes to darker blue. The red blocks always changes to darker red.   They do not change from blue to red or vice versa. This reflects that a neural network follows certain searching path during the training process. Based on such observations, we provide another view that can help users to visualize the updates of weight parameters during training phases. This is the trajectories of weight values over time.  

Figure~\ref{rwk4} presents 4 searching paths that the weight parameters in the first convolutional layer of our simplified LeNet-5 go through during 40 epochs(44,000 time steps). These trajectories are depicted by 3 arbitrary selected dimensions from the weight parameters. We use the selected weight values as x,y and z coordinates and color the path with time steps. The color of points  changes from gray to blue, then to red and finally to pink. All four paths are taken from the training that using the learning rate$=0.001$. An interesting observation is the relative straight head of all the 4 paths. The straight portion suggests that the neural network finds out a location that is close to a local minima in a very short time(approximately in the first 100-200 time steps). Then it takes the rest of training to climb down to the local minima like some Brownian motions. The trajectory in Figure\ref{rwk4}(b) is more interesting that the z coordinates look like several steps. We consider that the neural network wandering around places near one local minimum then finds a path so that it jumps near another local minimum.  

\begin{figure}
\includegraphics[width=0.95\linewidth]{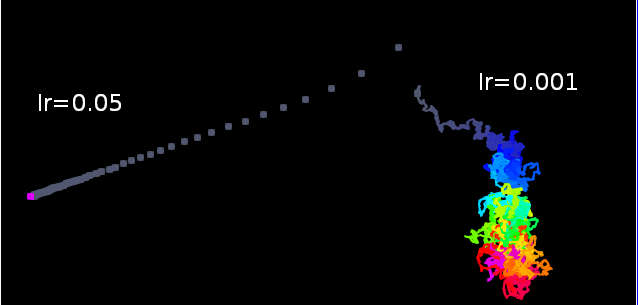}
\caption{Trajectories of two trainings with different learning rate. Left path: a network with big learning rate stops learning due to gradient vanish. Right path: A normal learning trajectory. }
\label{rwk_lrs}
\end{figure}

The trajectory view is a better tools that users can find out problematic learning rate. Figure\ref{rwk_lrs} shows the trajectory of two trainings with different learning rate. One trajectory is same as the above Figure\ref{rwk4}(a). The other is very straight. This is because we set the learning rate to be $0.05$ and the network stops learning due to gradient vanishing.    

\subsubsection{Visualizing Activations}
The weight-grid and weight trajectory provide qualitative visualizations about the neural network. Users can use them to see the updating of weight parameters during training phase. Our open framework provides distribution-grid and image-grid to visualize the feature maps or activations. Users can use these two grid as a quantitative measurement to decide if they have many redundant convolutional filters in their networks. 

\begin{figure*}[h]
\includegraphics[width=1.0\linewidth]{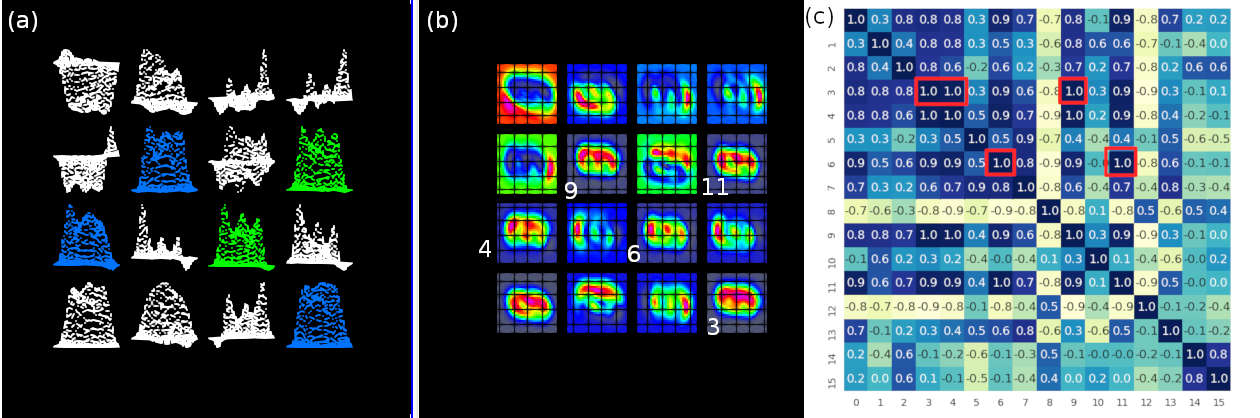}
\caption{The visualization of feature maps(activations) indicates there are redundant filters in the first convolutional layer in our simplified LeNet-5. (a)Distribution-grid of activations in the first convolutional layer. (b)Image-grid of the same activations. (c)Heatmap represents Pearson's Correlation Coefficients(PCC). In the heatmaps, darker blue stands for higher correlation values. Light blue or yellow are lower correlation values.}
\label{ah1_2}
\end{figure*}

Fig.\ref{ah1_2}(a) shows the distribution-grid of activations in the convolutional layer$_1$ (16 filters) in our simplified LeNet-5. Filters are numbered from 0 to 15 for the first convolutional layer. They starts from lower left to upper right. The subfigure(a) of Fig.\ref{ah1_2} uses blue color to mark filter 3,4 and 6 as group$_1$  and green color to mark filter 6 and 11 as group$_2$.  Our algorithm puts filters have high Pearson's Correlation Coefficient values into same groups. The image-grid(b) directly plots the feature maps. We add the filter index 3,4,9 for group$_1$ and 6,11 for group$_2$ to illustrate in this paper. The actual image-grid does not show the index numbers.  Although we will not quite agree with the image-grid that filter$_3$ is more similar to filter$_4$ and filter$_9$ than it is similar to filter$_11$, but users should be confident because the similarity is computed based on the Pearson's Correlation Coefficient. It is also a small fix for users to replace the PCC with other similarity metrics. The subfigure(c) shows the heatmap of Pearson's Correlation Coefficient values. We mark the PCC values of the above two groups. For the display space, we round the PCC value to the tenth precision. The real values are all around 0.97-0.99. The heatmap is provied only to illustrate how we render the colors in distribution-grids. They are not included in our framework. 

With our framework, users are able to visualize and identify redundant convolutional filters and use this information to adaptively prune neurons during the training process or compress a pretrained network. Pruning is shown in the next section.
 
\subsubsection{Neuron Pruning and in situ visualization}

\begin{figure*}[t]
\includegraphics[width=1.0\linewidth]{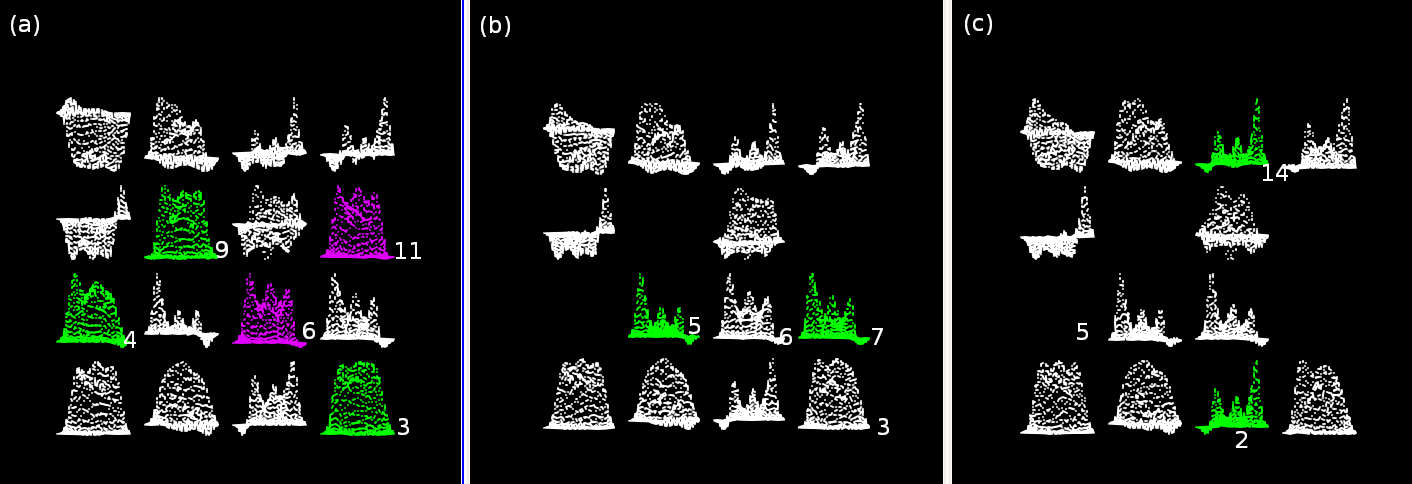}
\caption{(a)At step 1080, two groups are marked. Green group has\{3,4,9\} and pink group has\{6,11\}. (b)At step 1680,  redundant filters \{4,9\} and \{11\} are removed. \{3\} and \{6\} are kept.  A new group is marked that \{5,7\} are similar. (c)At step 2220, \{7\} is removed and \{5\} is kept. A new group is marked \{2,14\} are similar. After only 5 epochs, 16 filters reduced to 12.}
\label{cnvreduce}
\end{figure*}

%It has been shown~\cite{Srivastava:2014} that larger neural networks and networks that have been carelessly crafted are more likely to overfit. Then, automatically determining network topology has been a hot topic of research in the machine learning community. 
%These works mostly focus on the size and interconnection of layers and frequently employ brute force mechanisms. 
%As a complementary method of dropout, we propose our first approximation at dealing with this problem from an on-line pruning perspective.  
Figure~\ref{ah1_2} shows our open framework provide the distribution-grid and image-grid to suggest possible redundant filters in the first convolutional layer. Intuitively, they are slowing down the training and predicting process. They also have some probability to bring overfitting problems~\cite{Srivastava:2014}. Thus we want to remove such redundant neurons. 
  
The following is a simple proof about how we are able to remove the similar filters. Assume the network has two convoultional layers. They are layer$_1$ and layer$_2$. $x$ is an input batch of images. In layer$_1$, $w_1$ and $b_1$ are the weight and bias parameters in filter$_1$, $w_2$ and $b_2$ are the  weight and bias parameters of filter$_2$. $\sigma$ is the ReLU activation function($max(0,x)$). In layer$_2$, $w'_1$ and $w'_2$ are  weight parameters in filter$'$ . $b'$ is the bias of this filter$'$.   $a_1$ and $a_2$ are activations of filter$_1$ and filter$_2$ from layer$_1$. If $a_1$ and $a_2$ are similar, we can approximate $a'$ of filter$'$ in layer$_2$ with the following equations.
\begin{equation}
a_1 = \sigma(w_1 x + b_1) \approx a_2 = \sigma(w_2 x + b_2)
\end{equation}
\begin{equation}
a' = \sigma(w'_1 a_1 + w'_2 a_2 + b')
\end{equation}
\begin{equation}
a' \approx \sigma((w'_1 + w'_2) a_1 + b')
\end{equation}
The equation (3)  means we can simply add the corresponding $w'_1$  and $w'_2$ in layer$_2$ and remove one of filter$_1$ or filter$_2$ from layer$_1$ without incurring a big change in the loss.  Fig.\ref{cnvreduce} shows the distribution-grid at three time steps.  Our simplified LeNet-5 starts with 16 filters. At step 1080, our open framework marks two groups \{3,4,9\} and \{6,11\}. The green color indicates filters \{3,4,9\} in this group have high Pearson's Correlation Coefficient values. The pink color indicates filters \{6,11\} are similar. Subfigure(b) shows the distribution-grid at step 1680. Filter \{4,9\} and \{11\} are removed filter \{3\} and \{6\} are kept. The framework finds filter\{5,7\} are similar at this epoch. Subfigure(c) shows the distribution-grid at step 2220. Filter \{7\} is removed filter \{5\} is kept. The framework finds filter\{2,14\} are similar at this step and is ready to perform the pruning. Because our open framework uses the co-processing library of Paraview, we are able to visualize these distribution-grids in real time training of our simplified LeNet-5 network. With the prunining methods, we further reduced the 16 filter to 12 after 5 epochs(5500 steps) of training. 

\subsection{Pretrained VGG16 network}
In the second case study, we test the visualization methods on the VGG16 network. As more researchers today do not train their deep neural networks from scrtach, they take the pretrained parameters to fine-tune their own networks. Because the weight parameters are pretrained and fixed, we are more interested in visualizing the feature maps(activations) so that we can  find out the redundant filters. On the other hand, the pretrained parameters of VGG16 is quite big. The parameter file needs around 580 Megabytes storage. The network produces feature maps that contain 3 million pixels from the first convolutional layer at each time step. The activation data uses about 85 Megabytes per step,  which requests our methods to have good scalability in rendering the images.  Figure\ref{fig:teaser} illustrate the overview of image-grid(a),distribution-grid(b) and two zoomin feature maps(c and d).  In the following part of this  section, we discuss some design decisions and their visualization results.

Although the neural network is not trained from scratch, we want to visualize the pretrained VGG16 network's behavior on the original Imagenet dataset. The Imagenet training data contains $1,281,167$ RGB images. The validation data contains $50,000$ RGB images.  The first visualization we want to see is the image-grid.
    
\subsubsection{Visualization of image-grids}
\begin{figure}[h]
\includegraphics[width=0.95\linewidth]{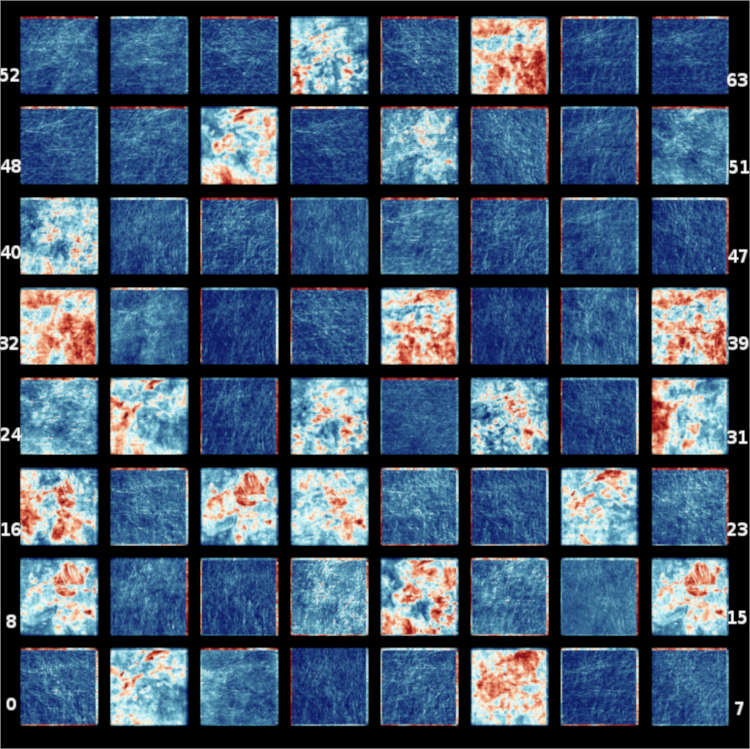}
\caption{Activations of the first training batch at VGG16's convolutional layer$_1$.}
\label{vgg-image-grid-0}
\end{figure}
\begin{figure}[!t]
\centering
\subfloat[Zoom in feature maps of filter \{8,15,18\}]{\includegraphics[width=1.0\linewidth]{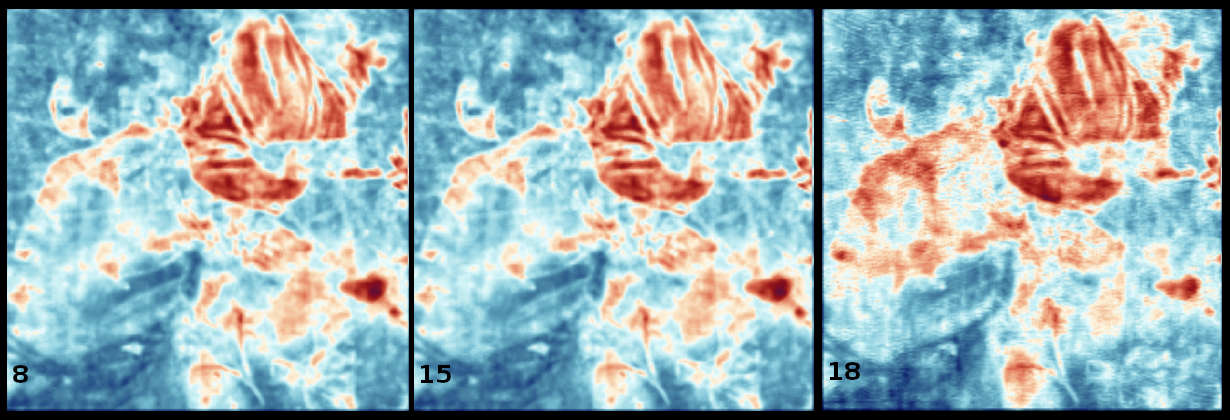}%
\label{vgg-image-zoomin}}
\hfil
\subfloat[Zoom in feature maps of filter \{52,53,54\}]{\includegraphics[width=1.0\linewidth]{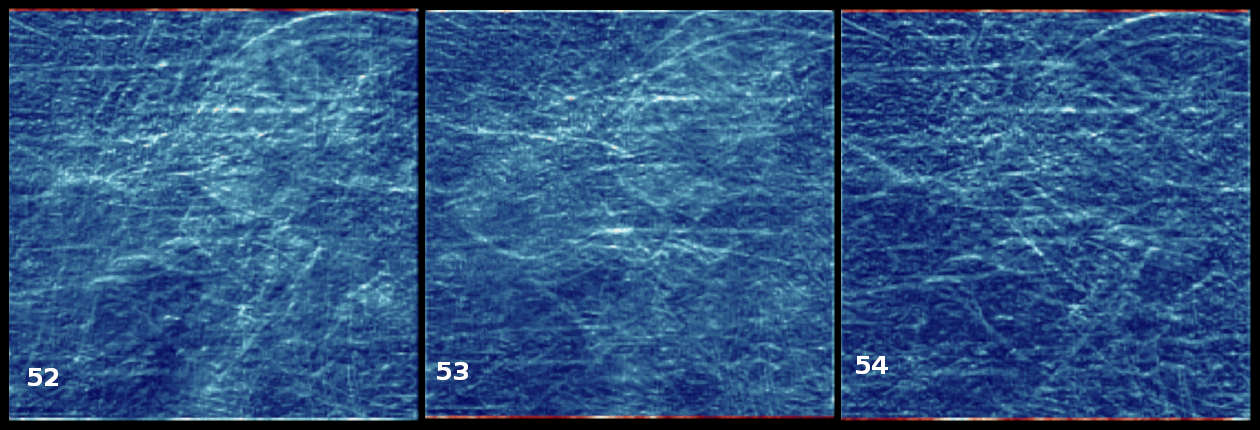}%
\label{vgg-image-texture}}
\caption{(a)In VGG16's convolutional layer$_1$,  the feature map of filter$_8$ is almost exactly the same as the feature map of filter$_15$. However, feature maps in subfigure(b) look like textures. It is difficult for users to decide whether they are similar or not. In this case, the distribution-grid is better for users to check similar feature maps.}
\label{fig_sim}
\end{figure}

The image-grid visualizes the feature maps in the same way as we did in the simplified LeNet-5 that classify the MNIST dataset. 
Figure\ref{vgg-image-grid-0} shows the feature map of the first training batch that produced by the convolutional layer$_1$. Because the number of pixels of the image-grid is more than 3 million, the ascii file format does not scale when we are rendering the image-grid. The VTK polygon data produced from the Catalyst co-processing pipeline helps to render the image-grid fast. The image-window starts from the lower left corner of this grid. Users can zoom in to check the feature maps. The zoomed in images in Figure\ref{vgg-image-zoomin} shows the feature maps for filter\{8,15,18\} are very similar. Thus users can find that these three filters are redundant quite easily.  However, some feature maps represent learned textures from the training batches. Figure\ref{vgg-image-texture} shows it is very difficult for users to check whether the texture learned by convolutional filter \{52,53,54\} are similar or not. In this case, the distribution-grid can help users to get better understand of the feature maps.

%\begin{figure}[h]
%\begin{subfigure}[b]{0.5\textwidth}
%\includegraphics[width=1.0\linewidth]{figures/vgg_image_zoomin}
%\caption{Zoom in feature maps of filter \{8,15,18\} }
%\label{vgg-image-zoomin}
%\end{subfigure}
%\begin{subfigure}[b]{0.5\textwidth}
%\includegraphics[width=1.0\linewidth]{figures/vgg_image_texture}
%\caption{Zoom in feature maps of filter \{52,53,54\}. }
%\label{vgg-image-texture}
%\end{subfigure}
%\caption{(a)In VGG16's convolutional layer$_1$,  the feature map of filter$_8$ is almost exactly the same as the feature map of filter$_15$. However, feature maps in subfigure(b) look like textures. It is difficult for users to decide whether they are similar or not. In this case, the distribution-grid is better for users to check similar feature maps.}
%\end{figure}

\begin{figure*}[t]
\centering
	\includegraphics[width=1.0\linewidth]{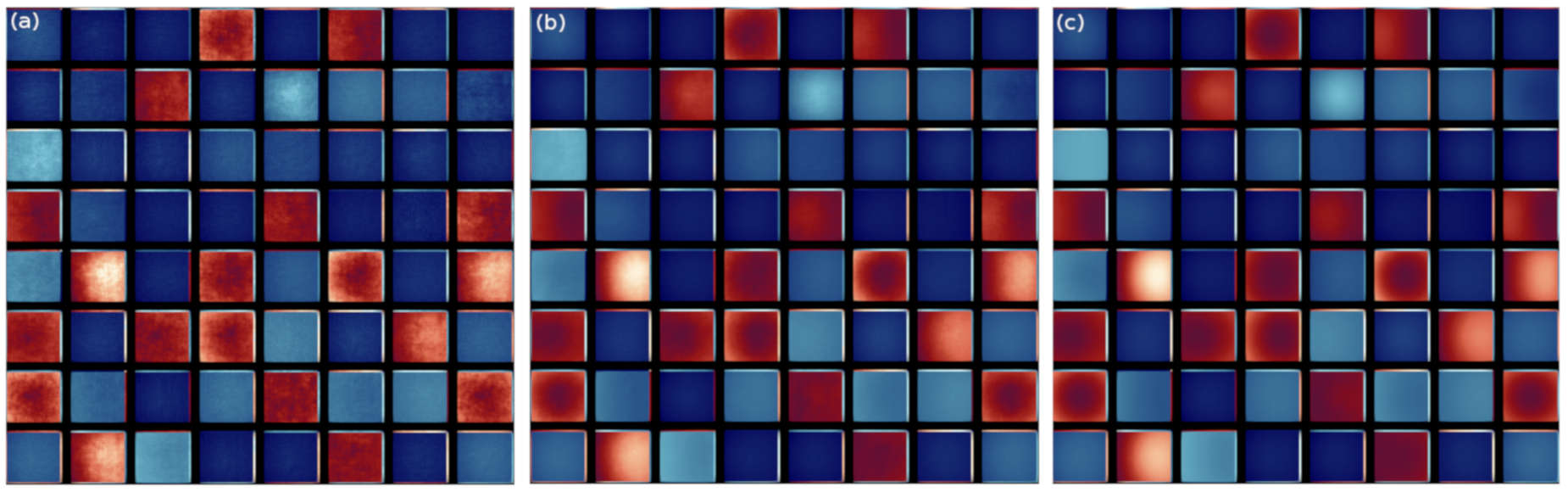}
	\caption{The image-grid of accumulated feature maps. (a)-(c) After 10,100 and 1000 training batches. }
	\label{vgg-image-sums}
\end{figure*}

Figure\ref{vgg-image-grid-0} visualizes the feature maps of one batch in the training dataset. As the VGG16 network learns more training batches, patterns displayed in each feature map  indicate various areas of interests correspond to each convolutional filter.  In Figure\ref{vgg-image-sums}, we show accumulated feature maps after 10,100 and 1000 batches. The light-colored regions in each feature map are smoothed through training. This reflects that each convolutional filter focuses on different part of the input images. The accumulated feature maps indicate their different focusing regions.

%\subfloat[The distribution-grid shows accumulated feature maps of the first training batch at VGG16's convolutional layer$_1$. Distribution-window of \{8,15,18\} are quite similar. \{32,61\} are similar. ]{\includegraphics[width=\linewidth]{figures/vgg_dist_0_short}
%\label{vgg-dist-grid-0}}
%\hfil
%\begin{figure*}[h]
%\centering
%	\includegraphics[width=\linewidth]{figures/vgg_dist_sums_short}
%	\caption{The distribution-grid of accumulated feature maps. (a)-(c) After 20,30 and 40 training batches. }
%	\label{vgg-dist-sums}
%\end{figure*}

\subsubsection{Visualization of distribution-grids}
\begin{figure}[h]
\centering
	\includegraphics[width=\linewidth]{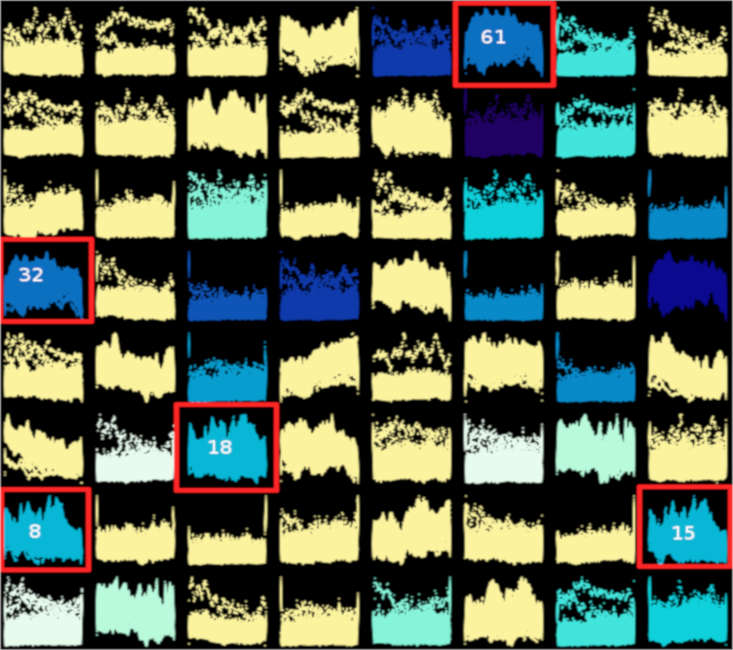}
	\caption{The distribution-grid shows accumulated feature maps of the first training batch at VGG16's convolutional layer$_1$. Distribution-window of \{8,15,18\} are quite similar. \{32,61\} are similar. }
	\label{vgg-dist-grid-0}
\end{figure}

In the distribution-grids, activations are flattened so users can see similarity between feature maps more easily. Figure\ref{vgg-dist-grid-0} display the distribution-grid after the pretrained VGG16 net consumes the first training batch. The two groups \{8,15,18\} and \{32,61\} marked in figure\ref{vgg-dist-grid-0} can be confirmed in the image-grid visualization(previous figure\ref{vgg-image-grid-0}).  

%An interesting observation from the distribution-grids is the distribution-window starts from patches like shown in figure\ref{vgg-dist-grid-0}. After some time steps, the distribution-windows display strips lithat shown in figure\ref{vgg-dist-sums}. This is analogous to the smooth effect we observe in the accumulated feature maps in figure\ref{vgg-image-sums}. Because the distribution-windows are computed by first flattening the accumulated feature maps then normalize them to the value between $[0,1]$, a lot of the details in the original feature maps are normalized to zero values. The strips represent the highest value after accumulation thus we can observe that filters are focusing in different regions. With the help of figure\ref{vgg-dist-sums}, we can answer the previous question about whether some texture feature maps are similar. We marked the strips in figure\ref{vgg-dist-sums}(a),(b) and (c). The feature map of \{56,57\} are more similar. Although feature map of \{8,15,18\} are similar as shown in the image-grid, we observe the feature maps of \{15,18\} are becoming more similar as the networks accumulates more training batches.

From the image-grid and distribution-grid, users can find out the similarities and decide if they want to compress and retrain the network. We did not experiment on merging similar filters and retraining this pretrained VGG16 network. Because the activations of similar approximate to each other, we expect the retraining of the pruned network can converge quickly in a few epochs.

\section{Conclusion}
In this paper, we present the in situ TensorView open framework. This open framework supports multiple output formats to visualize neural networks of different complexities. In situ TensorView leverages the scalability of Paraview Catalyst co-processing library to visualize and analyze the training, predicting and fine-tuning of large Covolutional Nueral Networks. It provides flexible views that visualize weight parameters and the activations of convolutional filters. These views include the weight-grids, weight trajectories, distribution-grids and image-grids. With these views, users can observe the dynamics of their networks during training and predicting phases. The statistical information provided by the distribution-grids can help users to decide whether some of the convolutional filters are redundant and guide them to prune or compress their networks.  

We present the visualizations in two case studies. They are a simplified LeNet-5 network and a pretrained VGG16 network. The experiments show that in situ TensorView can visualize both small and large convolutional neural networks. It can help users to visualize the networks in cases of training a network from scratch or fine-tuning some pretrained networks.

% or


%\appendices

%\section{}

%\ifCLASSOPTIONcompsoc
%\else
%\fi


\ifCLASSOPTIONcaptionsoff
  \newpage
\fi

\end{document}